\documentclass{article}

\usepackage[final]{corl_2019} % Uncomment for the camera-ready ``final'' version
\usepackage{graphicx}
\usepackage[utf8]{inputenc} % allow utf-8 input
\usepackage[T1]{fontenc}    % use 8-bit T1 fonts
\usepackage{hyperref}       % hyperlinks
\usepackage{url}            % simple URL typesetting
\usepackage{booktabs}       % professional-quality tables
\usepackage{amsfonts}       % blackboard math symbols
\usepackage{amsmath}
\usepackage{nicefrac}       % compact symbols for 1/2, etc.
\usepackage{microtype}      % microtypography
\usepackage{amssymb}
\usepackage{amsmath}
\usepackage{color}
\usepackage{microtype}
\usepackage{wrapfig}
\usepackage{algorithm}
\usepackage{algorithmic}
\usepackage{subcaption, bbm}

\title{Relay Policy Learning: Solving Long-Horizon Tasks via Imitation and Reinforcement Learning}

\author{
  Abhishek Gupta\thanks{Work done as an intern at Robotics at Google}\\
  Berkeley AI Research, Google\\
  \texttt{abhigupta@berkeley.edu} \\
  \And
  Vikash Kumar\\
  Google \\
  \texttt{vikashplus@gmail.com} \\
  \And
  Corey Lynch\\
  Google \\
  \texttt{coreylynch@google.com} \\
  \And
  Sergey Levine\\
  Berkeley AI Research, Google \\
  \texttt{svlevine@eecs.berkeley.edu} \\
  \And
  Karol Hausman\\
  Google \\
  \texttt{karolhausman@google.com}
}

\begin{document}
\maketitle

%===============================================================================

\begin{abstract}
We present relay policy learning, a method for imitation and reinforcement learning that can solve multi-stage, long-horizon robotic tasks. 
This general and universally-applicable, two-phase approach consists of an imitation learning stage that produces goal-conditioned hierarchical policies, and a reinforcement learning phase that finetunes these policies for task performance.
Our method, while not necessarily perfect at imitation learning, is very amenable to further improvement via environment interaction, allowing it to scale to challenging long-horizon tasks.
We simplify the long-horizon policy learning problem by using a novel data-relabeling algorithm for learning goal-conditioned hierarchical policies, where the low-level only acts for a \emph{fixed} number of steps, regardless of the goal achieved.
While we rely on demonstration data to bootstrap policy learning, we do not assume access to demonstrations of every specific tasks that is being solved, and instead leverage unstructured and unsegmented demonstrations of semantically meaningful behaviors that are not only less burdensome to provide, but also can greatly facilitate further improvement using reinforcement learning. 
We demonstrate the effectiveness of our method on a number of multi-stage, long-horizon manipulation tasks in a challenging kitchen simulation environment. Videos are available at \mbox{\url{https://relay-policy-learning.github.io/}}
\end{abstract}


%===============================================================================

\section{Introduction}

\begin{wrapfigure}{r}{0.4\textwidth}
    \centering
    \includegraphics[width=\linewidth]{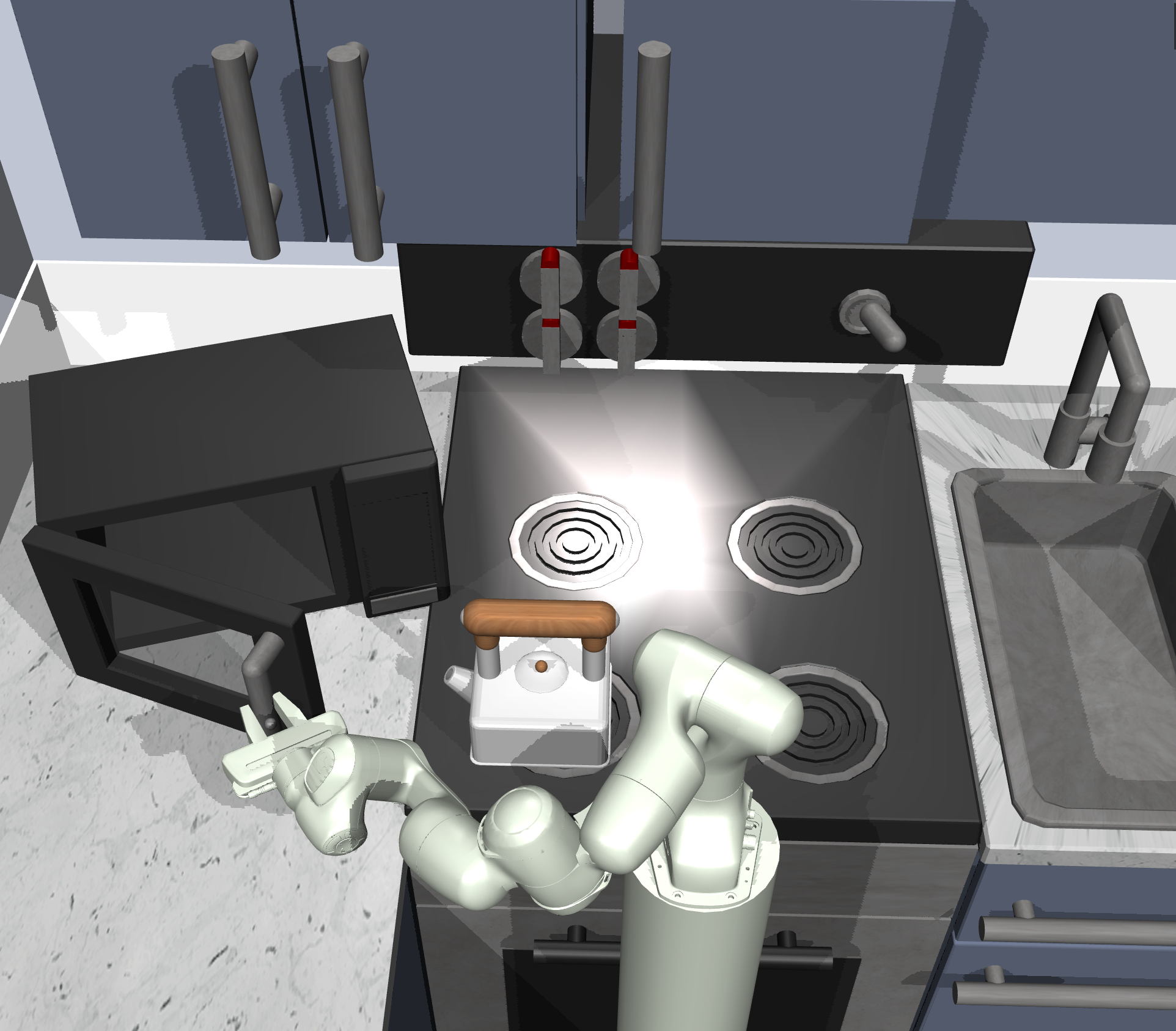}
    \caption{\footnotesize{RPL learns complex, long-horizon manipulation tasks}}
    \label{fig:frankakitchen}
    \vspace{-0.6cm}
\end{wrapfigure}

Recent years have seen reinforcement learning (RL) successfully applied to a number of robotics tasks such as in-hand manipulation~\cite{vikashICRA2016}, grasping~\cite{qtopt} and door opening~\cite{shaneDDPG}. However, these applications have been largely constrained to relatively simple short-horizon skills. Hierarchical reinforcement learning (HRL)~\cite{hrlreview} has been proposed as a potential solution that should scale to challenging long-horizon problems, by explicitly introducing temporal abstraction. However, HRL methods have traditionally struggled due to various practical challenges such as exploration~\cite{hdqn}, skill segmentation~\cite{options} and reward definition~\cite{diayn}. 
We can simplify the above-mentioned problems by utilizing extra supervision in the form of unstructured human demonstrations, in which case the question becomes: how should we best use this kind of demonstration data to make it easier to solve long-horizon robotics tasks?

This question is one focus area of hierarchical imitation learning (HIL), where solutions~\cite{SWIRL, DDO} typically try to achieve two goals: i) learn a temporal task abstraction, and ii) discover a meaningful segmentation of the demonstrations into subtasks.
These methods have not traditionally been tailored to further RL fine-tuning, making it challenging to apply them to a long-horizon setting, where pure imitation is very likely to fail. To address this need, we devise a simple and universally-applicable two-phase approach that in the first phase pre-trains hierarchical policies using demonstrations such that they can be easily fine-tuned using RL during the second phase.
In contrast to HRL methods, our method takes advantage of unstructured demonstrations to bootstrap further fine-tuning, and in contrast to conventional HIL methods, it does not focus on careful subtask segmentation, making the method simple, general and very amenable to further reinforcement fine-tuning. In particular, we show that we can develop an imitation and reinforcement learning approach that while not necessarily perfect at imitation learning, is very amenable to improvement via fine-tuning with reinforcement learning and that can be scaled to challenging long-horizon manipulation tasks.

What are the advantages of using such an algorithm? First, the approach is very general, in that it can be applied to any demonstration data, including easy to provide unsegmented, unstructured and undifferentiated demonstrations of meaningful behaviors. Second, our method does not require any explicit form of skill segmentation or subgoal definition, which otherwise would need to be learned or explicitly provided. Lastly, and most importantly, since our method ensures that every low-level trajectory is goal-conditioned (allowing for a simple reward specification) and of the same, limited length, it is very amenable to reinforcement fine-tuning, which allows for continuous policy improvement. We show that relay policy learning allows us to learn general, hierarchical, goal-conditioned policies that can solve long-horizon manipulation tasks in a challenging kitchen environment in simulation, while significantly outperforming hierarchical RL algorithms and imitation learning algorithms.

%===============================================================================

\section{Related Work}
\label{sec:citations}
Typical solutions for solving temporally extended tasks have been proposed under the HRL framework~\cite{hrlreview}. Solutions like the options framework~\cite{options, option-critic}, HAM~\cite{ham}, max-Q~\cite{maxq}, and feudal networks~\cite{feudal, FuN} present promising algorithmic frameworks for HRL. A particularly promising approach was proposed in~\citet{HIRO} and~\citet{HAC},
using goal conditioned policies at multiple layers of hierarchy for RL. Nevertheless, these algorithms still suffer from challenges in exploration and optimization (as also seen in our experimental comparison with ~\citet{HIRO}), which have limited their application to general robotic problems. In this work, we tackle these problems by using additional supervision in the form of unstructured, unsegmented human demonstrations. Our work builds on goal-conditioned RL~\cite{kaelbling, HER, TDM, UVFA}, which has been explored in the context of reward-free learning~\cite{RIG}, learning with sparse rewards~\cite{HER}, large scale generalizable imitation learning~\cite{LMP}, and hierarchical RL~\cite{HIRO}. We build on this principle to devise a general-purpose imitation and RL algorithm that uses data relabeling and bi-level goal conditioned policies to learn complex skills. 

There has a been a number of hierarchical imitation learning (HIL) approaches~\cite{intentionGAN, optionGAN, directedinfoGAIL, DDO, compile} that typically focus on extracting transition segments from the demonstrations. 
These methods aim to perform imitation learning by learning low-level primitives~\cite{DDO, compile} or latent conditioned policies~\cite{intentionGAN} which meaningfully segment the demonstrations. Traditionally, these approaches do not aim to and are not amenable to improving the learned primitives with subsequent RL, which is necessary as we move towards multi-task, challenging long-horizon problems where pure imitation might be insufficient.
%This is likely because it is hard to define a clear reward function and objective for continuous improvement of these primitives.
In this work, we specifically focus on utilizing both imitation and RL, and devise a method that does not explicitly try to segment out individual primitives into meaningful subtasks, but instead splits the demonstration data into fixed-length segments, amenable to fine-tuning with reinforcement learning. This allows us to leverage relabeling across different goals~\cite{kaelbling, HER, TDM, UVFA}. We introduce a novel form of goal relabeling and demonstrate its efficiency when applied to learning robust bi-level policies. A related idea is presented in~\citet{HIRL}, where the authors assume that an expert provides labelled and segmented demonstrations at both levels of the hierarchy, with an interactive expert for guiding RL. In contrast, we use a pool of unlabelled demonstrations and apply our method to learn a policy to achieve various desired goals, without needing interactive guidance or segmentation. 
Using imitation learning as a way to bootstrap RL has been previously leveraged by a number of deep RL algorithms~\cite{DAPG, yukeGAIL, ashvin}, where a flat imitation learning initialization is improved using reinforcement learning with additional auxiliary objectives. In this work, we show that we can learn hierarchical policies in a way that can be fine-tuned better than their flat counterparts.

\section{Preliminaries}
\label{sec:prelims}

\paragraph{Goal-conditioned reinforcement learning:} We define $\mathcal{M} = (S, A, P, r)$ to be a finite-horizon Markov decision process (MDP), where $S$ and $A$ are state and action spaces, $P(s_{t+1} \mid s_t, a_t)$ is a transition function, $r$ a reward function. The goal of RL is to find a policy $\pi(a | s)$ that maximizes expected reward over trajectories induced by the policy: $\mathbb{E}_{\pi}[\sum_{t=0}^T \gamma^t r_i(s_t, a_t)]$. To extend RL to multiple tasks, a goal-conditioned formulation (~\cite{kaelbling}) can be used to learn a policy $\pi(a | s, s_g)$ which maximizes the expected reward $r(a, s, s_g)$ with respect to a goal distribution $s_g \sim \mathcal{G}$ as follows: $\mathbb{E}_{s_g\sim\mathcal{G}} [\mathbb{E}_{\pi}[\sum_{t=0}^T \gamma^t r_i(s_t, a_t, s_g)]]$. 

\paragraph{Goal-conditioned imitation learning:} In typical imitation learning, instead of knowing the reward $r$, the agent has access to demonstrations $\mathcal{D}$ containing a set of trajectories $\mathcal{D} = \{\tau^i, \tau^j, \tau^k, ...\}$ of state-action pairs $\tau^i = \{s_0^i, a_0^i, \dots, s_T^i, a_T^i\}$. The goal is to learn a policy $\pi(a|s)$ that imitates the demonstrations. A common approach is to maximize the likelihood of actions in the demonstration, i.e. $\max E_{(s, a) \sim \mathcal{D}} \log\pi(a|s)$, referred to as behavior cloning (BC). When there are multiple demonstrated tasks, we consider a goal-conditioned imitation learning setup where the dataset of demonstrations $\mathcal{D}$ contains sequences that attempt to reach different goals $s_g^i, s_g^j, s_g^k, ...$. 
The objective is to learn a goal-conditioned policy $\pi(a|s, s_g)$ that is able to reach different goals $s_g$ by imitating the demonstrations. 

\section{Relay Policy Learning}
\label{sec:rpl}

In this section, we describe our proposed relay policy learning (RPL) algorithm, which leverages unstructured demonstrations and reinforcement learning to solve challenging long-horizon tasks. 
Our approach consists of two phases: relay imitation learning (RIL), followed by relay reinforcement fine-tuning (RRF) described in Sec.~\ref{sec:ril} and~\ref{sec:finetuning} respectively. While RIL by itself is not able to solve the most challenging tasks that we consider, it provides a very effective initialization for fine-tuning. 

\begin{figure}[!h]
    \centering
    \includegraphics[width=0.8\textwidth]{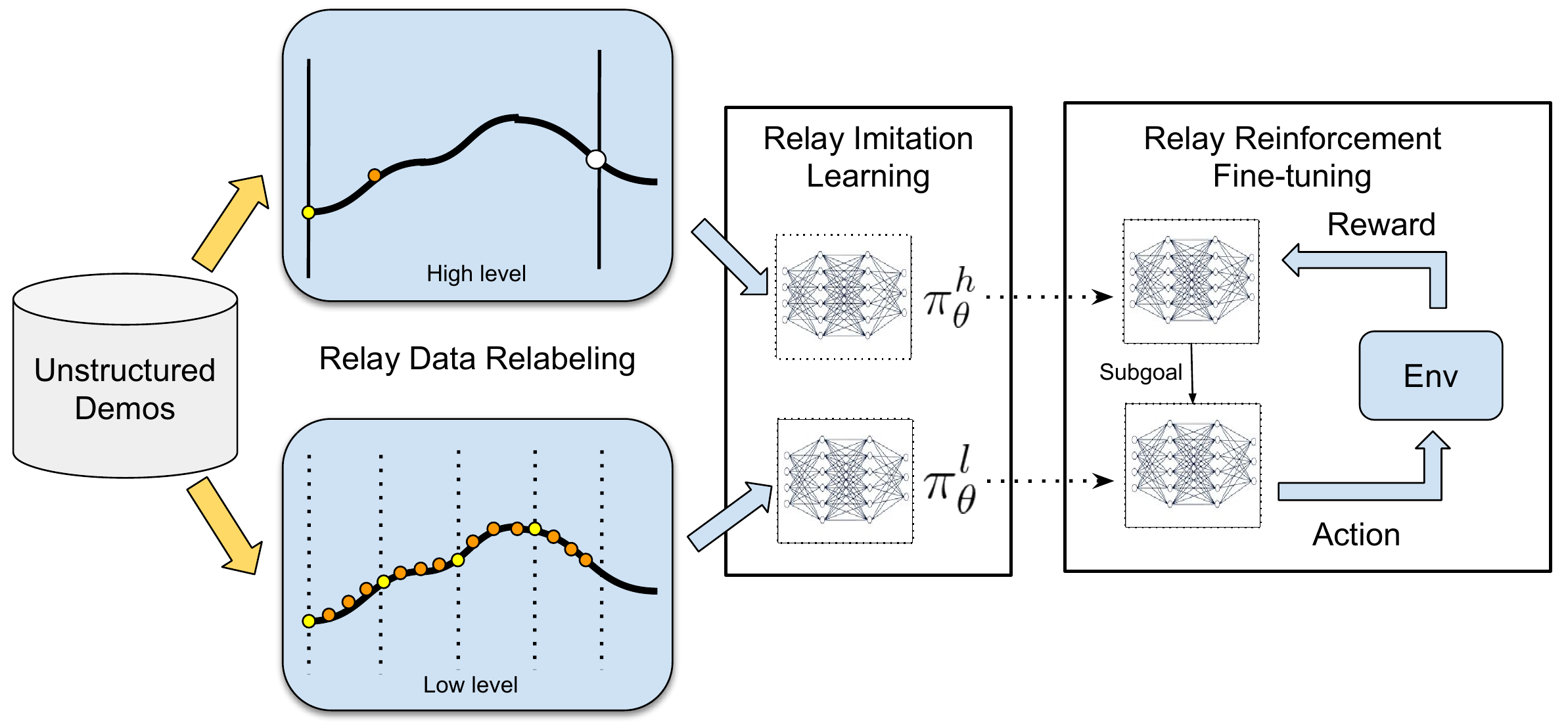}
    \caption{\footnotesize{\textbf{Relay policy learning:} the algorithm starts with relabelling unstructured demonstrations at both the high and the low level of the hierarchical policy and then uses them to perform relay imitation learning. 
    This provides a good policy initialization for subsequent relay reinforcement fine-tuning. 
    We demonstrate that learning such simple goal-conditioned policies at both levels from demonstrations using relay data relabeling, combined with relay reinforcement fine-tuning allows us to learn complex manipulation tasks.}}
    \label{fig:mainfig}
\end{figure}

\subsection{Relay Policy Architecture}
\label{sec:rpa}

\begin{wrapfigure}{r}{0.55\textwidth}
    \centering
    \includegraphics[width=.5\textwidth]{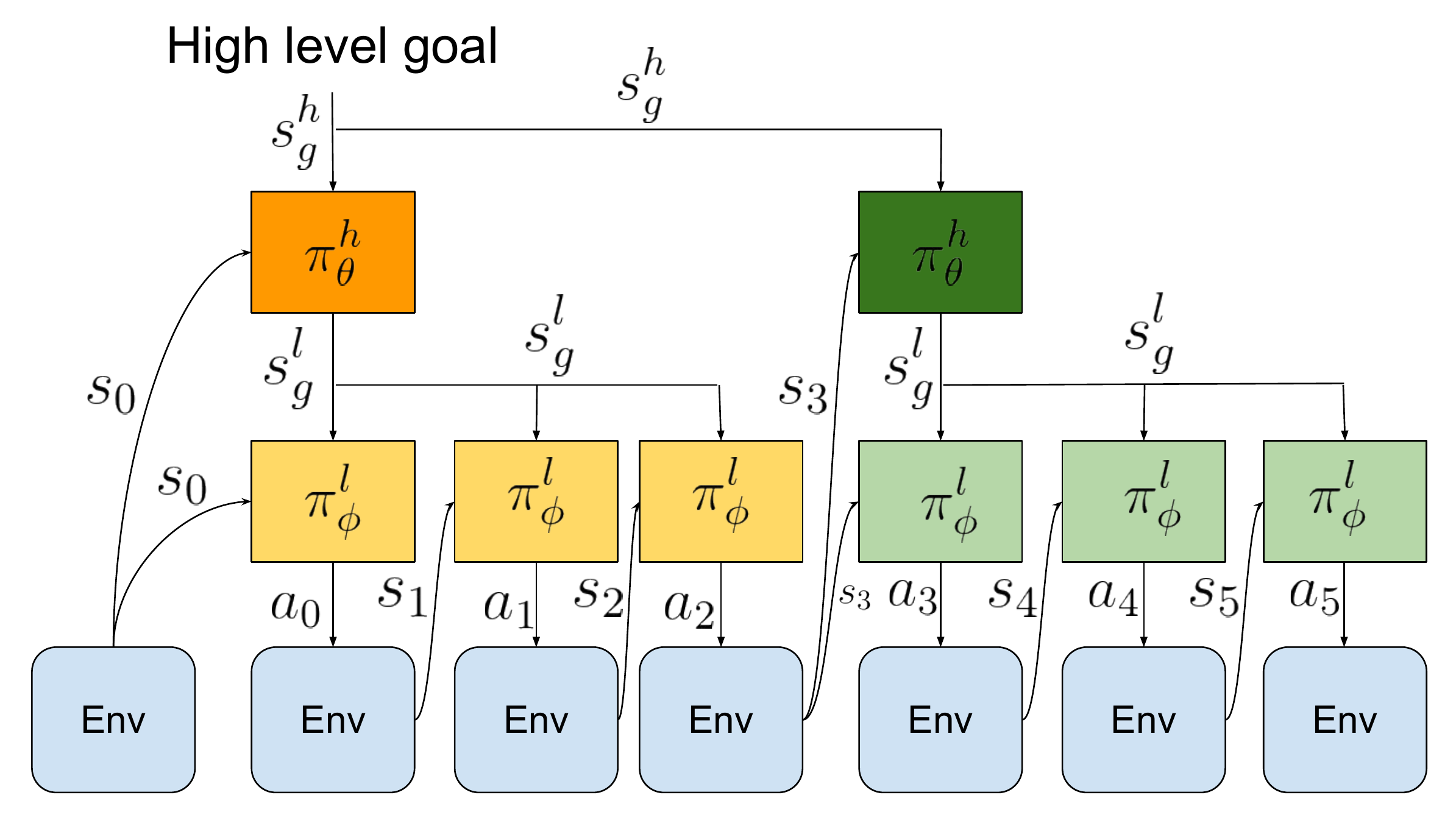}
    \caption{\footnotesize{\textbf{Relay policy architecture:} A high level goal setter $\pi_{\theta}$ takes high level goal $s_g^h$ and sets goals $s_g^l$ for a lower level policy $\pi_{\phi}$, which acts for a fixed time horizon before resampling $s_g^l$}}
    \label{fig:hrl_policy}
    \vspace{-0.5cm}
\end{wrapfigure}

We first introduce our bi-level hierarchical policy architecture (shown in Fig~\ref{fig:hrl_policy}), which enables us to leverage temporal abstraction. 
This architecture consists of a high-level goal-setting policy and a low-level subgoal-conditioned policy, which together generate an environment action for a given state. 
The high-level policy $\pi_{\theta}^{h}(s_g^l|s_t, s_g^h)$ takes the current state $s_t$ and a long-term high-level goal $s_g^h$ and produces a subgoal $s_g^l \in \mathcal{S}$ which is then ingested by a low-level policy $\pi_{\phi}^{l}(a|s_t, s_g^l)$. 
The low-level policy takes the current state $s_t$, and the subgoal $s_g^{l}$ commanded by the high-level policy and outputs an action $a_t$, which is executed in the environment. 

Importantly, the goal setting policy $\pi_{\theta}^{h}$ makes a decision every $H$ time steps (set to $30$ in our experiments), with each of its subgoals being kept constant during that period for the low-level policy, while the low-level policy $\pi_{\phi}^{l}$ operates at every single time-step. 
This provides temporal abstraction, since the high level policy operates at a coarser resolution than the low-level policy. 
This policy architecture, while inspired by goal-conditioned HRL algorithms~\cite{HIRO}, requires a novel learning algorithm to be applicable in the context of imitation learning, which we describe in Sec.~\ref{sec:ril}. Given a high-level goal $s_g^h$, $\pi_{\theta}^{h}$ samples a subgoal $s_{g_0}^l$, which is passed to $\pi_{\theta}^{l}$ to generate action $a_0$. For the subsequent $H$ steps, the goal produced by $\pi_{\theta}^{h}$ is kept fixed, while $\pi_{\theta}^{l}$ generates an action $a_t$ at every time step. 

\begin{figure*}
\noindent
\begin{minipage}{0.495\textwidth}
\begin{algorithm}[H]
\caption{\footnotesize Relay Policy Learning}
\label{alg:rpl}
\begin{algorithmic}[1]
{\footnotesize
\REQUIRE Unstructured pool of demonstrations $D = \{\tau_0, \tau_1,...\tau_N\}$
\STATE Relabel goals in demonstration trajectories using Algorithm~\ref{alg:ril_relabellow},~\ref{alg:ril_relabelhigh} to extract $D_l, D_h$ 
\STATE \textbf{Relay Imitation Learning:} Train $\pi_{\theta}^h$ and $\pi_{\phi}^l$ using Eqn~\ref{eqn:maxlikelihood}
\WHILE{not done}
    \STATE Collect on-policy experience with $\pi_{\theta}^h$ and $\pi_{\phi}^l$ for high  level goals different $s_g^h$
    \STATE [Optional] Relabel this experience (Sec.~\ref{sec:finetuning}), and add to $D_l$, $D_h$
    \STATE Update the policy via policy gradient update using Eqn~\ref{eqn:lowlevel},~\ref{eqn:highlevel}. 
\ENDWHILE
\STATE Distill fine-tuned policies into a single multi-goal policy 
}
\end{algorithmic}
\end{algorithm}
\end{minipage}
\noindent
\hfill
\begin{minipage}{0.49\textwidth}
\begin{algorithm}[H]
\caption{\footnotesize Relay data relabeling for RIL low level}
\label{alg:ril_relabellow}
\begin{algorithmic}[1]
{\footnotesize
\REQUIRE Demonstrations $D = \{\tau_0, \tau_1,...\tau_N\}$
\FOR{$n=1...N$}
    \FOR{$t=1...t_n$}
        \FOR{$w=1...W_l$}
          \STATE Add $(s_t^n, a_t^n, s_{t + w}^n)$ to $D_l$
        \ENDFOR
    \ENDFOR
 \ENDFOR
}
\end{algorithmic}
\end{algorithm}
\vspace{-0.8cm}
\begin{algorithm}[H]
\caption{\footnotesize Relay data relabeling for RIL high level}
\label{alg:ril_relabelhigh}
\begin{algorithmic}[1]
{\footnotesize
\REQUIRE Demonstrations $D = \{\tau_0, \tau_1,...\tau_N\}$
\FOR{$n=1...N$}
    \FOR{$t=1...t_n$}
        \FOR{$w=1...W_h$}
          \STATE Add $(s_t^n, s_{t + \min(w, W_l)}^n, s_{t + w}^n)$ to $D_h$
        \ENDFOR
    \ENDFOR
 \ENDFOR
}
\end{algorithmic}
\end{algorithm}
\end{minipage}
\end{figure*}

\subsection{Relay Imitation Learning}
\label{sec:ril}

Our problem setting assumes access to a pool of unstructured, unlabeled ``play" demonstrations (\citet{LMP}) $\mathcal{D}$, corresponding to demonstrations of meaningful activities provided by the user, without any particular task in mind, e.g. opening cabinet doors, playing with different objects, or simply tidying up the scene. 
We do not assume that this demonstration data actually accomplishes any of the final task goals that we will need to solve at test-time, though we do need to assume that the test-time goals come from the same distribution of goals as those accomplished in the demonstration data.
In order to take the most advantage of such data, we initialize our policy with our proposed relay imitation learning (RIL) algorithm. RIL is a simple imitation learning procedure that builds on the goal relabeling scheme described in \citet{LMP} for the hierarchical setting, resulting in improved handling of multi-task generalization and compounding error. RIL assumes access to the pool of demonstrations consisting of $N$ trajectories $\mathcal{D} = \{\tau^i, \tau^j, \tau^k, ...\}$, where each trajectory consists of state-action pairs $\tau^i = \{s_0^i, a_0^i, \dots, s_T^i, a_T^i\}$. 
Importantly, these demonstrations can be attempting to reach a variety of different high level goals $s_g^h$, but we do not require these goals to be specified explicitly. To learn the relay policy from these demonstrations, we construct a low-level dataset $\mathcal{D}_l$, and a high-level dataset $\mathcal{D}_h$ from these demonstrations via ``relay data relabeling", which is described below, and use them to learn $\pi_{\theta}^{h}$ and $\pi_{\theta}^{l}$ via supervised learning at multiple levels. 

We construct the low-level dataset by iterating through the pool of demonstrations and relabeling them using our relay data relabelling algorithm. First, we choose a window size $W_{l}$ and generate state-goal-action tuples for $\mathcal{D}_l$, $(s, s_g^l, a)$ by goal-relabeling within a sliding window along the demonstrations, as described in detail below and in Algorithms~\ref{alg:ril_relabellow},~\ref{alg:ril_relabelhigh}. 
The key idea behind relay data relabeling is to consider all states that are actually reached along a demonstration trajectory within $W_{l}$ time steps from any state $s_t$ to be goals reachable from the state $s_t$ by executing action $a_t$. 
This allows us to label all states $s_{t+1}, ...., s_{t + W_{l}}$ along a valid demonstration trajectory as potential goals that are reached from state $s_t$, when taking action $a_t$. 
We repeat this process for all states $s_t$ along all the demonstration trajectories being considered. 
This procedure ensures that the low-level policy is proficient at reaching a variety of goals from different states, which is crucial when the low-level policy is being commanded potentially different goals generated by the high-level policy. 

We employ a similar procedure for the high level, generating the high-level state-goal-action dataset $\mathcal{D}_h$. However, the actions at the high level are subgoal states that are provided to the low-level policy, so they must be chosen as states along the demonstration trajectories. We start by choosing a high-level window size $W_h$, which encompasses the high-level goals we would like to eventually reach. 
We then generate state-goal-action tuples for $\mathcal{D}_h$, via relay data relabeling within the high-level window being considered, as described in Algorithm~\ref{alg:ril_relabellow},~\ref{alg:ril_relabelhigh}. 
We also label all states $s_{t+1}, ...., s_{t + W_{h}}$ along a valid trajectory as potential high-level goals that are reached from state $s_t$ by the high level policy, but we set the high-level action for a goal $j$ steps ahead $s_{t + j}$, as $s_{t + \min(W_l,  j)}$ choosing a sufficiently distant subgoal as the high-level action.  

Given these relay-data-relabeled datasets, we train $\pi_{\theta}^{l}$ and $\pi_{\theta}^{h}$ by maximizing the likelihood of the actions taken given the corresponding states and goals:
\begin{align}
\label{eqn:maxlikelihood}
&\max_{\phi, \theta} \mathbb{E}_{(s, a, s_g^l) \sim D_l}[\log \pi_{\phi}(a|s, s_g^l)] + \mathbb{E}_{(s, s_g^l, s_g^h) \sim D_h}[\log \pi_{\theta}(s_g^l|s, s_g^h)].
\end{align}
This procedure gives us an initialization for both the low-level and the high-level policies, without the requirement for any explicit goal labeling from a human demonstrator. As we show in our experiments, this bi-level initialization is significantly more amenable to RRF than learning the high level from scratch as described in~\cite{diayn, HIRO, LMP}, and allows us to avoid the expensive goal labeling that is required in~\cite{HIRL}. 
Relay data relabeling not only allows us to learn hierarchical policies without explicit labels, but also provides algorithmic improvements to imitation learning: (i) it generates more data through the relay-data-relabelling augmentation, and (ii) it improves generalization since it is trained on a large variety of goals. 

\subsection{Relay Reinforcement Fine-tuning}
\label{sec:finetuning}
The procedure described in Sec.~\ref{sec:ril} allows us to extract an effective policy initialization via relay imitation learning. 
However, this policy is often unable to perform well across all temporally extended tasks, due to the well-known compounding errors stemming from imitation learning~\cite{dagger}. 
Reinforcement learning provides a solution to this challenge, by enabling continuous improvement of the learned policy directly from experience. We can use RL to improve RIL policies via fine-tuning on different tasks. We employ a goal-conditioned HRL algorithm that is a variant of natural policy gradient (NPG) with adaptive step size~\cite{TRPO}, where both the high-level and the low-level goal-conditioned policies $\pi_{\theta}^h$ and $\pi_{\phi}^l$ are being trained with policy gradient in a decoupled optimization.

Given a low-level goal-reaching reward function $r_{l}(s_t, a_t, s_g^l)$, we can optimize the low-level policy by simply augmenting the state of the agent with the goal commanded by the high-level policy and then optimizing the policy to effectively reach the commanded goals by maximizing the sum of its rewards. 
For the high-level policy, given a high-level goal-reaching reward function $r_{h}(s_t, g_t, s_g^h)$, we can optimize it by running a similar goal-conditioned policy gradient optimization to maximize the sum of high-level rewards obtained by commanding the current low-level policy. 

To effectively incorporate demonstrations into this reinforcement learning procedure, we leverage our method via: (1) initializing both $\pi_{\theta}^{l}$ and $\pi_{\theta}^{h}$ with the policies learned via RIL, and (2) encouraging policies at both levels to stay close to the behavior shown in the demonstrations. 
To incorporate (2), we augment the NPG objective with a max-likelihood objective that ensures that policies at both levels take actions that are consistent with the relabeled demonstration pools $D_{l}$ and $D_{h}$ from Section~\ref{sec:ril}, as described in Eqn~\ref{eqn:lowlevel} and~\ref{eqn:highlevel}:  
\begin{align}
    \label{eqn:lowlevel}
    &\nabla_{\phi} J_l = \mathbb{E}[\nabla_{\phi}\log\pi_{\phi}^{l}(a|s,s_g^l) \sum_t r_{l}(s_t, a_t, s_g^l)] + \lambda_l\mathbb{E}_{(s,a,s_g^l) \sim \mathcal{D}_{l}}[\nabla_{\phi}\log\pi_{\phi}^{l}(a|s,s_g^l)]\\
    \label{eqn:highlevel}
    &\nabla_{\theta} J_h = \mathbb{E}[\nabla_{\theta}\log\pi_{\theta}^{h}(s_g^l|s,s_g^h) \sum_t r_{h}(s_t, s_g^l, s_g^h)] + \lambda_h\mathbb{E}_{(s,s_g^l,s_g^h) \sim \mathcal{D}_{h}}[\nabla_{\theta}\log\pi_{\theta}^{h}(s_g^l|s,s_g^h)].
\end{align}
While a similar objective has been described in~\cite{DAPG, ashvin}, it is yet to be explored in the hierarchical, goal-conditioned scenarios, which makes a significant difference as indicated in our experiments.

In addition, since we are learning goal-conditioned policies at both the low and high level, we can leverage relay data relabeling as described in Sec.~\ref{sec:ril} to also enable the use of off-policy data for fine-tuning.
Suppose that at a particular iteration $i$, we sampled $N$ trajectories according to the scheme proposed in Sec.~\ref{sec:rpa}. 
While these trajectories did not necessarily reach the goals that were originally commanded, and therefore cannot be considered optimal for those goals, they do end up reaching the \emph{actual} states visited along the trajectory. 
Thus, they can be considered as optimal when the goals that they were intended for are relabeled to states along the trajectory via relay data relabeling described in Algorithm~\ref{alg:ril_relabellow}, ~\ref{alg:ril_relabelhigh}. This scheme generates a low-level dataset $\mathcal{D}_{l}^i$ and a high level dataset $\mathcal{D}_{h}^i$ by relabeling the trajectories sampled at iteration $i$. 
Since these are considered ``optimal'' for reaching goals along the trajectory, they can be added to the buffer of demonstrations $D_{l}$ and $D_{h}$, thereby contributing to the objective described in Eqn~\ref{eqn:lowlevel} and Eqn~\ref{eqn:highlevel} and allowing us to leverage off-policy data during RRF. 
We experiment with three variants of the fine-tuning update in our experimental evaluation: IRIL-RPL (fine-tuning with Eqn~\ref{eqn:lowlevel},~\ref{eqn:highlevel} and iterative relay data relabeling to incorporate off-policy data as described above), DAPG-RPL (fine-tuning the policy with the update in Eqn~\ref{eqn:lowlevel},~\ref{eqn:highlevel} without the off-policy addition) and NPG-RPL (fine-tuning the policy with the update in Eqn~\ref{eqn:lowlevel},~\ref{eqn:highlevel}, without the off-policy addition or the second maximum likelihood term). 
The overall method is described in Algorithm~\ref{alg:rpl}.

As described in ~\citet{dnc}, it is often difficult to learn multiple tasks together with on-policy policy gradient methods, because of high variance and conflicting gradients. 
To circumvent these challenges, we use RPL to fine-tune on a number  of different high level goals individually, and then \emph{distill} all of the learned behaviors into a single policy as described in ~\citet{policydistillation}. 
This allows us to learn a single policy capable of achieving multiple high level goals, without dealing with the challenges of multi-task optimization.

\section{Experimental Results}
\label{sec:result}

Our experiments aim to answer the following questions: (1) Does RIL improve imitation learning with unstructured and unlabelled demonstrations? (2) Is RIL more amenable to RL fine-tuning than its flat, non-hierarchical alternatives? (3) Can we use RPL to accomplish long-horizon manipulation tasks? Videos and further experimental details are available at \mbox{\url{https://relay-policy-learning.github.io/}}

\begin{figure}[!h]
    \centering
    \begin{subfigure}[b]{.24\linewidth}
        \centering\includegraphics[trim=0 2cm 0 0, clip, width=\linewidth]{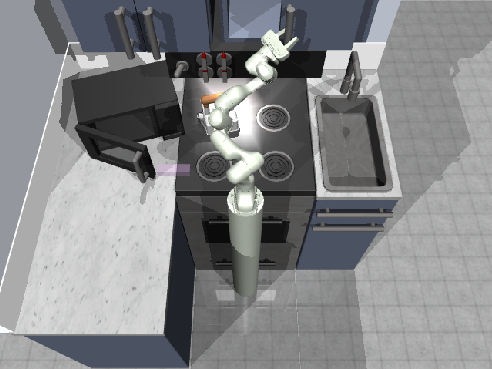}
        \caption{}
    \end{subfigure}
        \begin{subfigure}[b]{.24\linewidth}
        \centering\includegraphics[trim=0 2cm 0 0, clip, width=\linewidth]{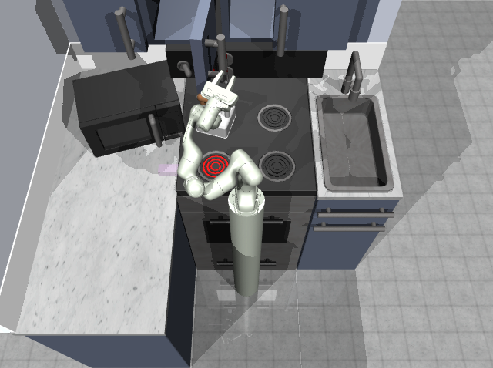}
        \caption{}
    \end{subfigure}
    \begin{subfigure}[b]{.24\linewidth}
        \centering\includegraphics[trim=0 2cm 0 0, clip, width=\linewidth]{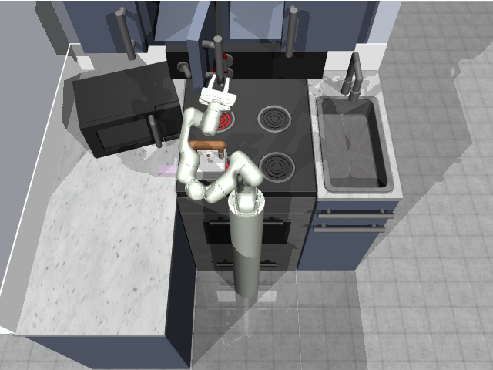}
        \caption{}
    \end{subfigure}
    \begin{subfigure}[b]{.24\linewidth}
        \centering\includegraphics[trim=0 2cm 0 0, clip, width=\linewidth]{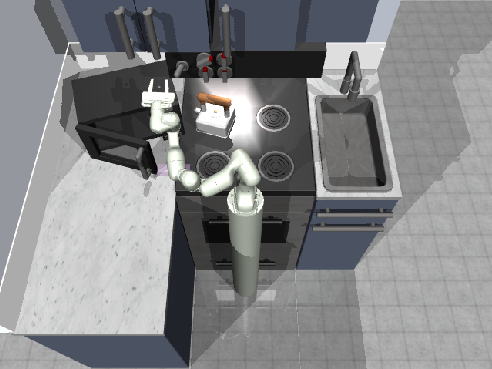}
        \caption{}
    \end{subfigure}
    \caption{\footnotesize{Examples of compound goals in the kitchen environment. Each goal has different elements manipulated, requiring multiple stages to solve: (a) microwave, kettle, light, slider, (b) kettle, burner, slider, cabinet, (c) burner, top burner, slide hinge, (d) kettle, microwave, top burner, lights}}
    \label{fig:frankatasks}
\end{figure}

\paragraph{Environment Setup}
\label{sec:expsetup}
To evaluate our algorithm, we utilize a challenging robotic manipulation environment modeled in MuJoCo, shown in Fig.~\ref{fig:frankakitchen}. The environment consists of a 9 DoF position-controlled Franka robot interacting with a kitchen scene that includes an openable microwave, four turnable oven burners, an oven light switch, a freely movable kettle, two hinged cabinets, and a sliding cabinet door. 
We consider reaching different goals in the environment, as shown in Fig.~\ref{fig:frankatasks}, each of which may require manipulating many different components. 
For instance, in Fig.~\ref{fig:frankatasks} (a), the robot must open the microwave, move the kettle, turn on the light, and slide open the cabinet. 
While the goals we consider are temporally extended, the setup is fully general. We collect a set of unstructured and unsegmented human demonstrations described in Sec.~\ref{sec:ril}, using the PUPPET MuJoCo VR system~\cite{puppet}. 
We provide the algorithm with 400 sequences containing various unstructured demonstrations that each manipulate four different elements of the scene in sequence.

\paragraph{Evaluation and Comparisons}
Since each of our tasks consist of compound goals that involve manipulating four elements in the environment, we evaluate policies based on the number of steps that they complete out of four, which we refer to as step-completion score.  A step is completed when the corresponding element in the scene is moved to within $\epsilon$ distance of its desired position.

We compare variants of our RPL algorithm to a number of ablations and baselines, including prior algorithms for imitation learning combined with RL and methods that learn from scratch. Among algorithms which utilize imitation learning combined with RL, we compare with several methods that utilize flat behavior cloning with additional finetuning. Specifically, we compare with (1) flat goal-conditioned behavior cloning followed by finetuning (\emph{BC}), (2) flat goal-conditioned behavior cloning trained with data relabeling followed by finetuning (\emph{GCBC})~\cite{LMP}, and variants of these algorithms that augment the \emph{BC} and \emph{GCBC} fine-tuning with losses as described in~\citet{DAPG} - (3) \emph{DAPG-BC} and (4) \emph{DAPG-GCBC}. We also compare RPL to (5) hierarchical imitation learning + finetuning with an oracle segmentation scheme, which performs hierarchical goal conditioned imitation learning by using a hand-specified oracle to segment the demonstrations for imitation learning, followed by RRF style fine-tuning. Details of this scheme can be found in Appendix 3. For comparisons with methods that learn from scratch we compare with (6) an on-policy variant of \emph{HIRO}~\cite{HIRO} trained from scratch with natural policy gradient~\cite{TRPO} instead of Q-learning and (7)  a baseline (\emph{Pre-train low level}) that learns low-level primitives from the demonstration data, and learns the high-level goal-setting policy from scratch with RL. The last baseline is representative of a class of HIL algorithms~\cite{intentionGAN, optionGAN, compile}, which are difficult to fine-tune because it is not clear how to provide rewards for improving low-level primitives. Lastly, we compare RPL with a baseline (7) (\emph{Nearest Neighbor}) which uses a nearest neighbor strategy to choose the demonstration which has the achieved goal closest to the commanded goal and subsequently executes actions open-loop.

\subsection{Relay Imitation Learning from Unstructured Demonstrations}
We start by aiming to understand whether RIL improves imitation learning over standard methods. We compare the step-wise completion scores averaged over 17 different compound goals with RIL as compared to flat BC variants. We find that, while none of the variants are able to achieve near-perfect completion scores via just imitation, the average stepwise completion score is higher for RIL as compared to both flat variants (see Table~\ref{fig:bc_comparison}, first row). 
Additionally, we find that the flat policy with data augmentation via relabeling performs better than without relabeling. 
When we analyze the proportion of compound goals that are actually fully achieved (see Table~\ref{fig:bc_comparison}, bottom row), RIL shows significant improvement over other methods. 
This indicates that, even for imitation learning, we see benefits from introducing the simple RIL scheme described in Sec.~\ref{sec:ril}. 

\begin{table}[!h]
\begin{center}
\begin{tabular}{ |c|c|c|c| } 
 \hline
  & \textbf{RIL (ours)} & GCBC relabeling & GCBC no relabeling\\ 
 \hline
 Success Rate (\%)& \textbf{21.7} & 8.8 & 7.6 \\ 
 \hline
 Average Step Completion (of 4) & \textbf{2.4} $\pm$ \textbf{1.13} & $2.2 \pm 0.95$ & $1.78 \pm 1.0$ \\ 
 \hline
\end{tabular}
\end{center}
\caption{\footnotesize{Comparison of  RIL to goal-conditioned behavior cloning with and without relabeling in terms success and step-completion rate averaged across 17 tasks. RIL outperforms the non-hierarchical methods}}
\label{fig:bc_comparison}
\vspace{-0.5cm}
\end{table}

\subsection{Relay Reinforcement Fine-tuning of Imitation Learning Policies}

\begin{figure}[!h]
    \centering
    \includegraphics[width=0.45\textwidth]{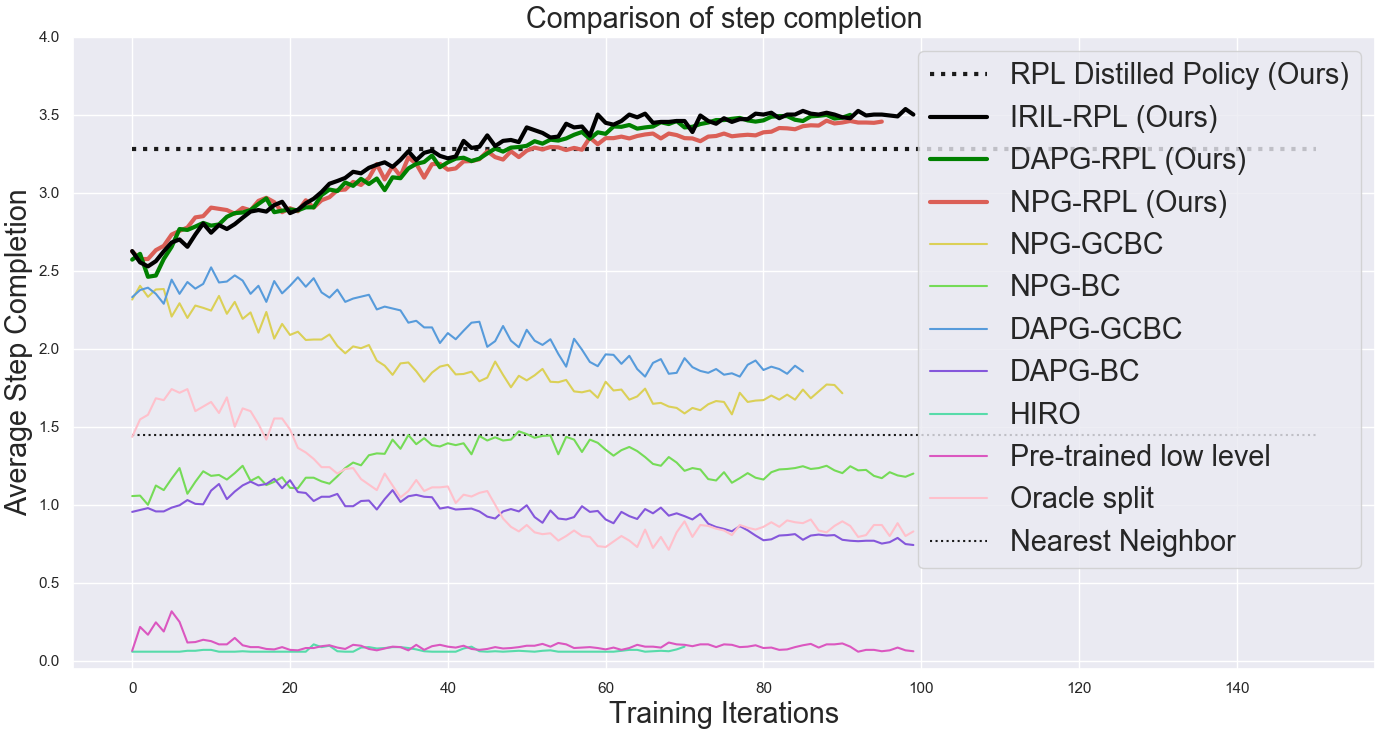}
    \includegraphics[width=0.45\textwidth]{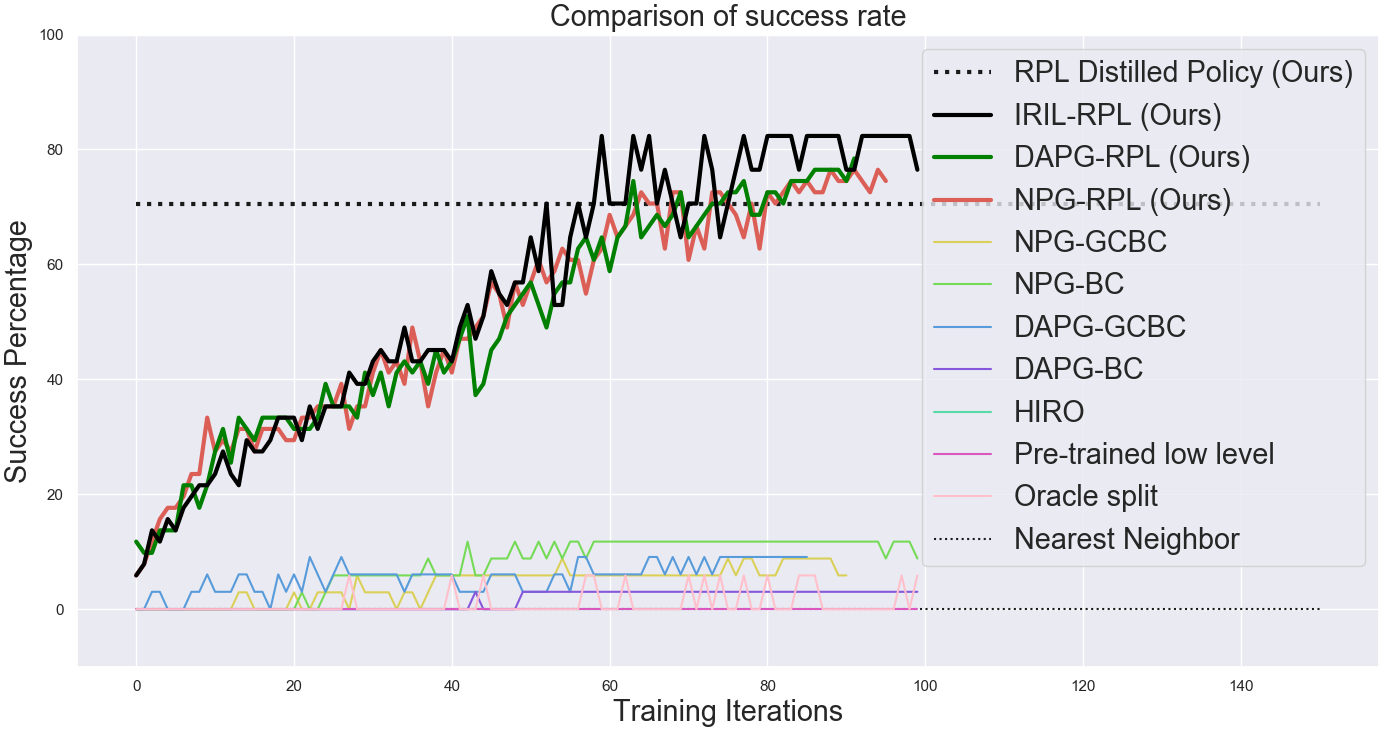}
    \caption{\footnotesize{Comparison of the RPL algorithm with a number of baselines averaged over 17 compound goals and 2 (baseline methods) or 3 (our approach) random seeds.  
    Fine-tuning with all three variants of our method outperforms fine-tuning using flat policies. 
    RIL initialization at both levels improves the performance over HIRO~\cite{HIRO} and over learning only the high-level policy from scratch. 
    If we use policy distillation, we are able to get a successful, multi-task goal-conditioned policy.}}
    \label{fig:rl_finetuning}
    \vspace{-0.5cm}
\end{figure}

Although pure RIL does succeed at times, its performance is still relatively poor. In this section, we study the degree to which RIL-based policies are amenable to further reinforcement fine-tuning. Performing reinforcement fine-tuning individually on 17 different compound goals seen in the demonstrations, we observe a significant improvement in the average success rate and stepwise completion scores over all the baselines when using any of the variants of RPL (see Fig.~\ref{fig:rl_finetuning}). 
In our experiments, we found that it was sufficient to fine-tune the low-level policy, although we could also fine-tune both levels, at the cost of more non-stationarity.
Although the large majority of the benefit is from RRF, we find a slight additional improvement from the DAPG-RPL and IRIL-RPL schemes, indicating that including the effect of the demonstrations throughout the process helps. 

When compared with HRL algorithms that learn from scratch (on-policy HIRO~\cite{HIRO}), we observe that RPL is able to learn much faster and reach a much higher success rate, showing the benefit of demonstrations. Additionally, we notice better fine-tuning performance when we compare RPL with flat-policy fine-tuning. 
This can be attributed to the fact that the credit assignment and reward specification problems are much easier for the relay policies, as compared to fine-tuning flat policies, where a sparse reward is rarely obtained.
The RPL method also outperforms the pre-train-low-level baseline, which we hypothesize is because we are not able to search very effectively in the goal space without further guidance. We also see a significant benefit over using the oracle scheme described in Appendix 3, since the segments become longer making the exploration problem more challenging. The comparison with the nearest neighbor baseline also suggests that there is a significant benefit from actually learning a \emph{closed-loop} policy rather than using an open-loop policy. While plots in Fig.~\ref{fig:rl_finetuning} show the average over various goals when fine-tuned individually, we can also distill the fine-tuned policies into a single, multi-task policy, as described in Sec.~\ref{fig:rl_finetuning}, that is able to solve almost all of the compound goals that were fine-tuned. 
While the success rate drops slightly, this gives us a single multi-task policy that can achieve multiple temporally-extended goals (Fig~\ref{fig:rl_finetuning}). 

\subsection{Ablations and Analysis}
\begin{figure}[!h]
    \vspace{-0.3cm}
    \centering
    \includegraphics[width=0.45\textwidth]{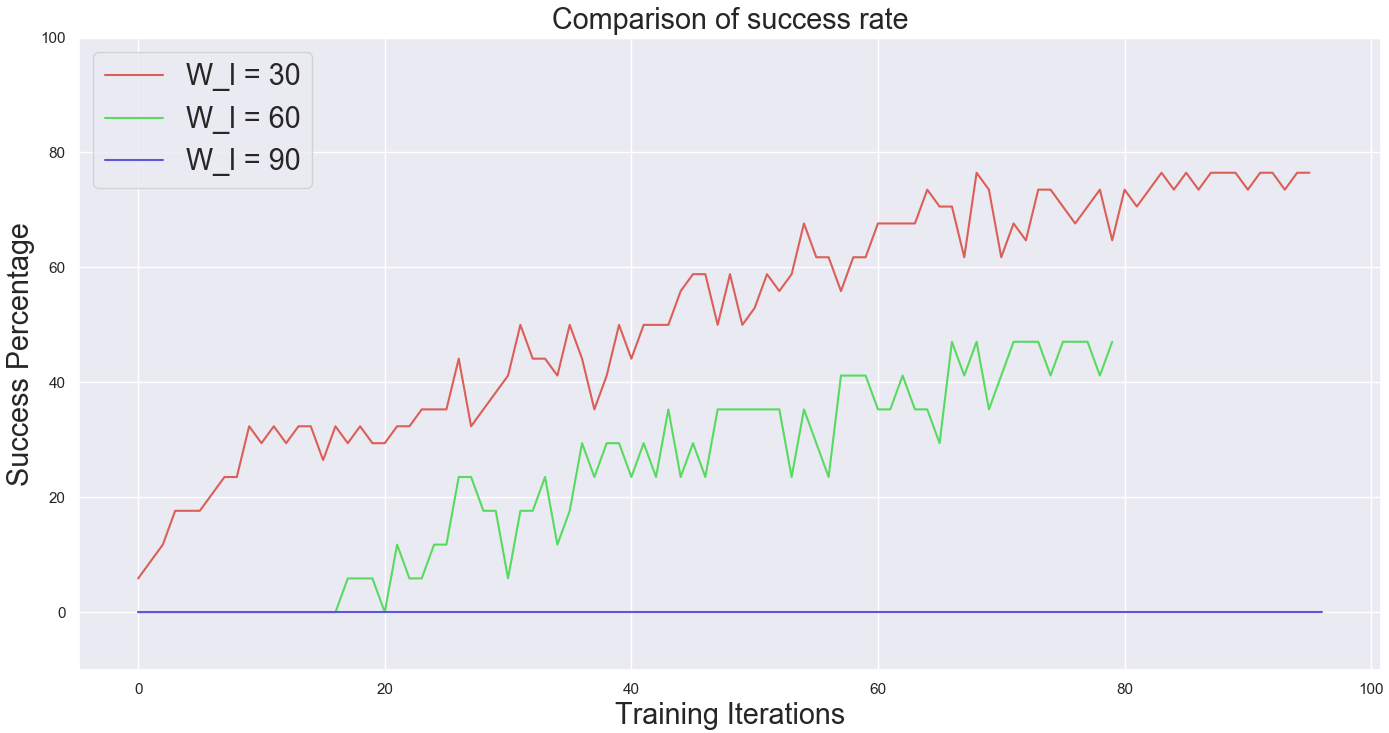}
    \includegraphics[width=0.45\textwidth]{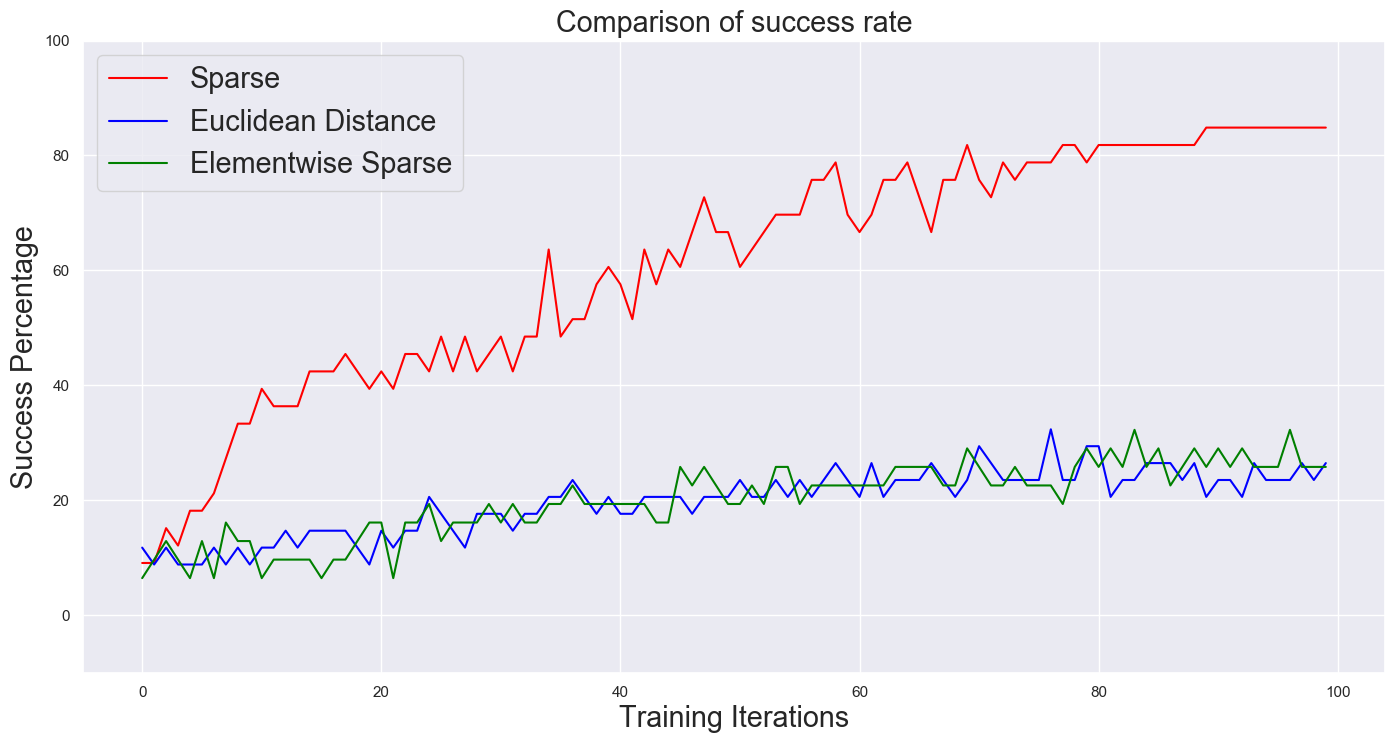}
    \caption{\footnotesize{\textbf{Left:} Role of low level window size in RPL. As the window size increases, imitation learning and fine-tuning become less effective. \textbf{Right:} Role of fine-tuning reward function in RPL. We see that the sparse reward function is most effective once exploration is sufficiently directed.}}
    \label{fig:ablations}
    \vspace{-0.3cm}
\end{figure}

To understand design choices, we consider the role of using different window sizes for RPL as well as the role of reward functions during fine-tuning. In Fig~\ref{fig:ablations} (left), we observe that the window size for RPL plays a major role in algorithm performance. 
As window size increases, both imitation learning and fine-tuning performance decreases since the behaviors are now more temporally extended.

Next, we consider the role of the chosen reward function in fine-tuning with RRF. We evaluate the relative performance of using different types of rewards for fine-tuning - sparse reward, euclidean distance, element-wise reward (refer to Appendix A for details). When each is used as a goal conditioned reward for fine-tuning the low-level, sparse reward works much better. This indicates that when exploration is sufficient, sparse reward functions are less prone to local optima than alternatives.

\section{Conclusion and Future Work}
\label{sec:conclusion}
We proposed relay policy learning, a method for solving long-horizon, multi-stage tasks by leveraging unstructured demonstrations to bootstrap a hierarchical learning procedure. 
We showed that we can learn a single policy capable of achieving multiple compound goals, each requiring temporally extended reasoning. 
In addition, we demonstrated that RPL significantly outperforms other baselines that utilize hierarchical RL from scratch, as well as imitation learning algorithms. 

In future work, we hope to tackle the problem of generalization to longer sequences and study extrapolation beyond the demonstration data. 
We also hope to extend our method to work with off-policy RL algorithms, so as to further improve data-efficiency and enable real world learning on a physical robot.  

\section{Acknowledgements}
We would like to thank Byron David for his help in improving our kitchen environment and simulation. We would also like to thank Suraj Nair, Chelsea Finn, Ofir Nachum, Michael Ahn, Anusha Nagabandi, Dibya Ghosh for fruitful discussions. We also thank Robotics at Google for a wonderful research atmosphere.

\bibliography{example}  % .bib

\appendix

\section{Experimental Details}
We use feed-forward MLPs for all our policies, with two layer neural networks with 256 units each and ReLu nonlinearities used for both the high-level policy $\pi_{\theta}$ and the low-level policy $\pi_{\phi}$ in all methods. Flat baselines use the same architecture as well and additional experimentation with the architecture did not yield substantially different results. We train all imitation learning algorithms with the ADAM optimizer using a batch size of $128$ and a learning rate of $0.005$. We choose $W_l$ to be $30$ and $W_h$ to be $260$ in all experiments. Our ablations suggest that the larger the window, the harder the learning problem becomes for both imitation and RL fine-tuning. 

For reinforcement learning, we utilize a variant of Trust Region Policy Optimization (TRPO). We fine-tune on 17 different compound goals individually, with a path length of $280$ for every compound goal, and the low-level horizon set to $30$. We use $100$ trajectories in each iteration of on-policy fine-tuning, with a discount of $0.995$. When using variants of augmenting the policy gradient objective with demonstrations, we experimented with different weights $\lambda_h$ and $\lambda_l$, but we found $0.0001$ to work well. We use a batch size of a $100$ trajectories per iteration, and fairly standard parameters for truncated natural policy gradient based on \url{https://github.com/aravindr93/mjrl}

The simulation environment has a 30-dimensional state space which consists of positions of the arm and the objects in the scene. The action space is 9 dimensional with 7 DoF for the arm and 2 DoF for the gripper. The actions are represented as the joint velocity. 

\section{Reward Function Details}
For the comparisons detailed in Section 5.3, the reward functions used for sparse, euclidean and element-wise sparse reward functions are detailed below, with $\epsilon$ set to $0.3$. For all our experimental results in Fig 5, we use the sparse reward variant as the reward function for fine-tuning. 

\begin{equation}
    R_{\text{sparse}}(s, g) = \mathbbm{1}(\|s - g\|_2 < \epsilon)
\end{equation}

\begin{equation}
    R_{\text{euclidean}}(s, g) = -\|s - g\|_2
\end{equation}

\begin{equation}
    R_{\text{elementwise sparse}}(s, g) = \sum_{\text{idx} \in \text{element indices}} \mathbbm{1}(\|s[\text{idx}] - g[\text{idx}]\|_2 < \epsilon)
\end{equation}

In the element-wise sparse reward case, $\text{idx}$ is selected to be the indices of state corresponding to different distinct elements of the scene such as the microwave, stove burners, light switch, sliding cabinet, hinge cabinets and so on. The robot arm is excluded from these indices. 

\section{Oracle Baseline Details}
\begin{figure}[!h]
    \centering
    \includegraphics[width=\linewidth]{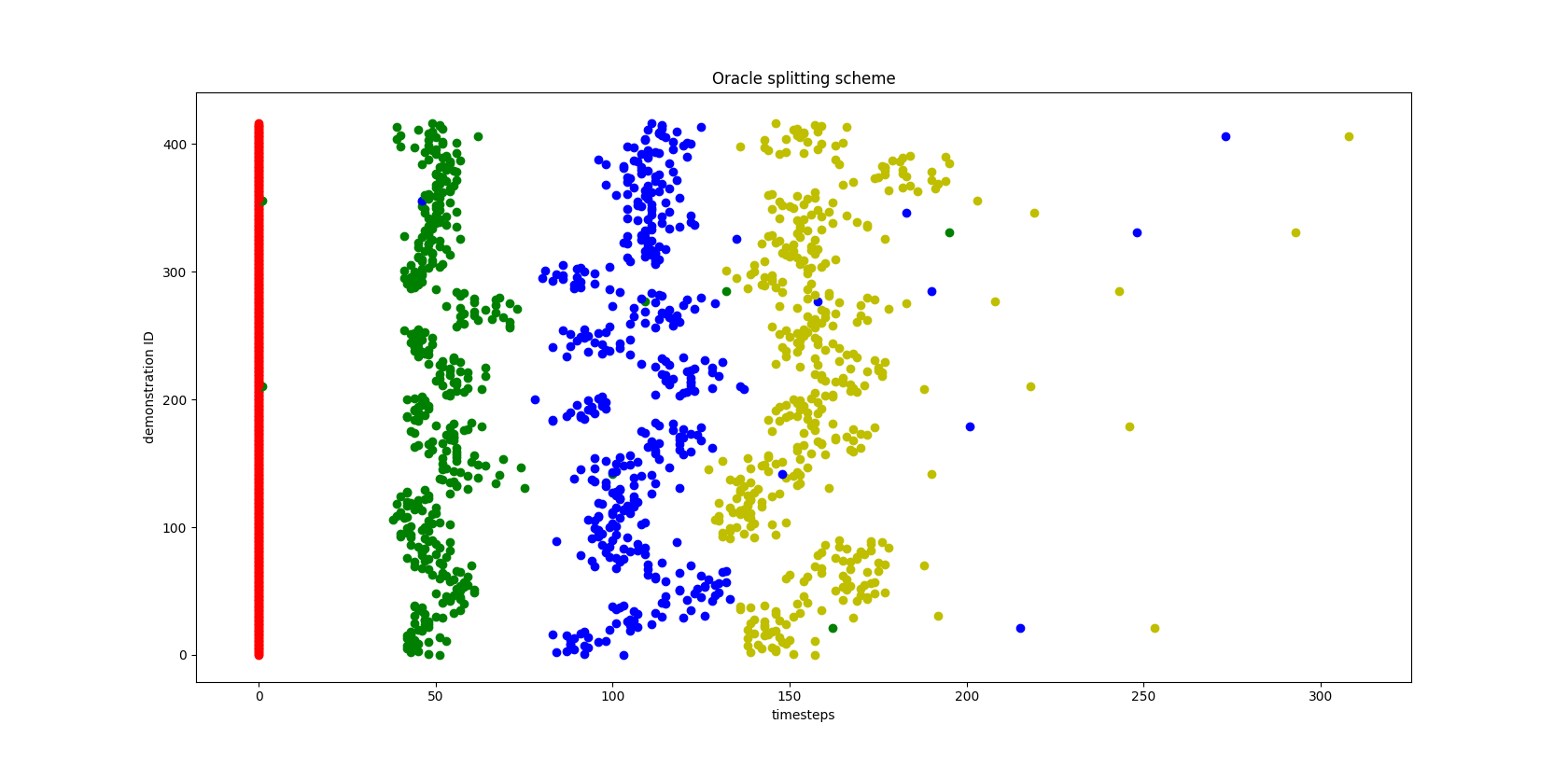}
    \caption{Splits generated by the oracle segmentation scheme. Each color corresponds to a different split and different demonstrations as plotted as different rows along the y-axis, with time-steps along the x-axis. We see that the split of demonstrations is fairly variable in time-steps. This makes the imitation learning and fine-tuning quite challenging.}
    \label{fig:split}
\end{figure}

For the oracle comparison described in Section 5, a hand-designed scheme is used to segment the demonstration into segments corresponding to semantically meaningful components, thereby generating variable sized windows rather than fixed length ones. Specifically, we split a segment any time one of [microwave, kettle, light switch, burners, slide cabinet, hinge cabinet] is moved more than $\epsilon = 0.3$. This leads to a variable segment generation scheme, which generates splits that is shown in Fig~\ref{fig:split}.

Segments generated in this fashion can then be used for imitation learning both the low-level and high-level policies. Specifically, the actions for the high level policies are chosen to be the states at which the segments are broken, and the low level is trained via goal conditioned behavior cloning with those states set as goals.

\section{Visualization of Learned Behaviors}
We show example visualizations of several successful learned behaviors for compound tasks, and some failed behaviors to better understand the the method. These can be best appreciated by viewing the accompanying videos on the supplementary website \mbox{\url{https://relay-policy-learning.github.io}}.

\subsection{Successful cases} 
\begin{figure}[!h]
    \centering
    \includegraphics[width=0.19\textwidth]{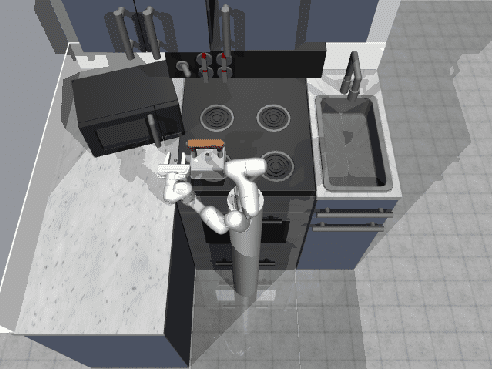}
    \includegraphics[width=0.19\textwidth]{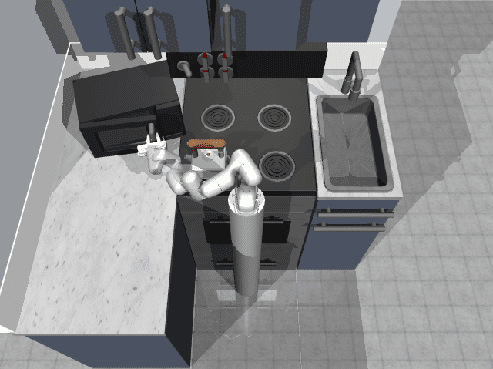}
    \includegraphics[width=0.19\textwidth]{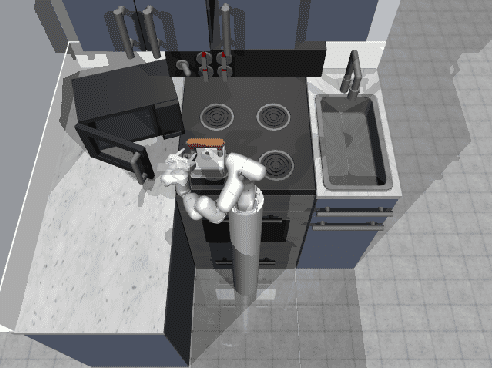}
    \includegraphics[width=0.19\textwidth]{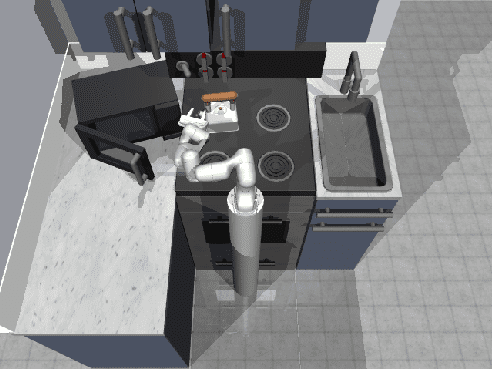}
    \includegraphics[width=0.19\textwidth]{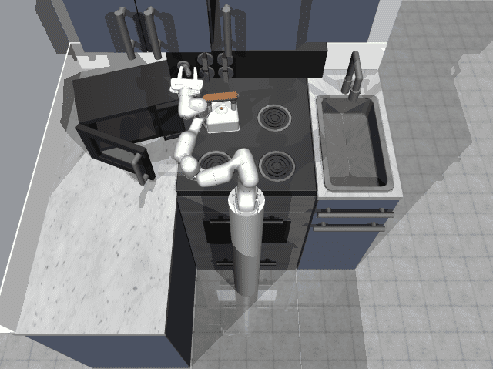}\\
    \includegraphics[width=0.19\textwidth]{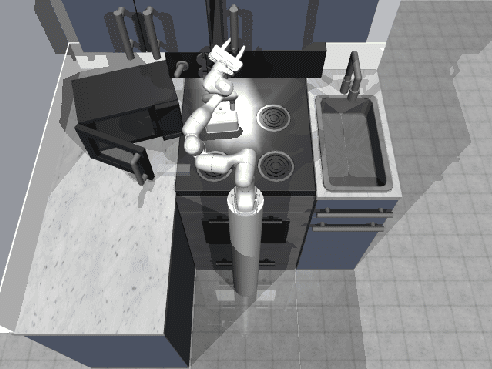}
    \includegraphics[width=0.19\textwidth]{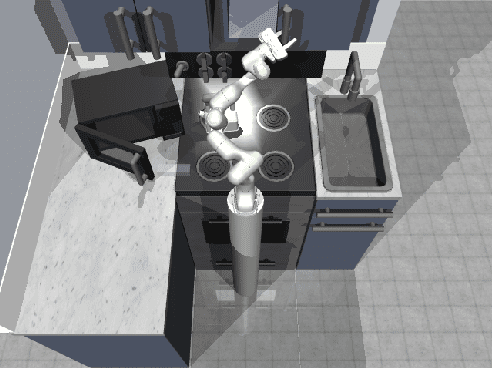}
    \includegraphics[width=0.19\textwidth]{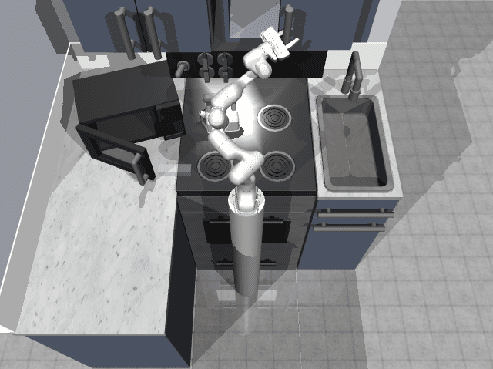}
    \includegraphics[width=0.19\textwidth]{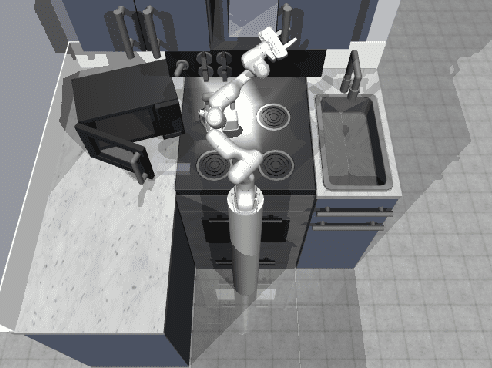}
    \includegraphics[width=0.19\textwidth]{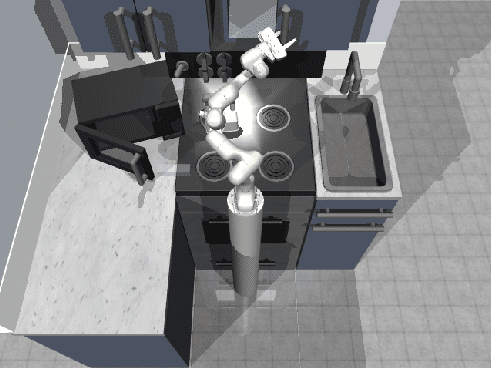}
    \caption{Visualization of successful learned behavior for opening microwave, moving kettle, turning on light switch, sliding the slider}
\end{figure}

\begin{figure} [H]
    \centering
    \includegraphics[width=0.19\textwidth]{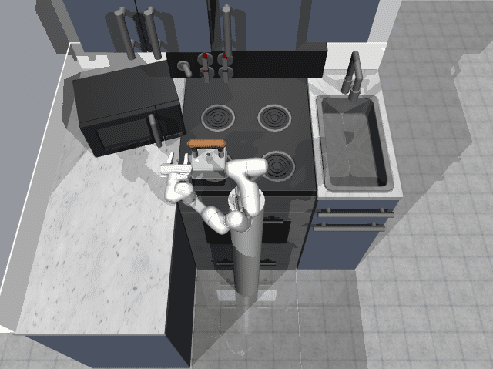}
    \includegraphics[width=0.19\textwidth]{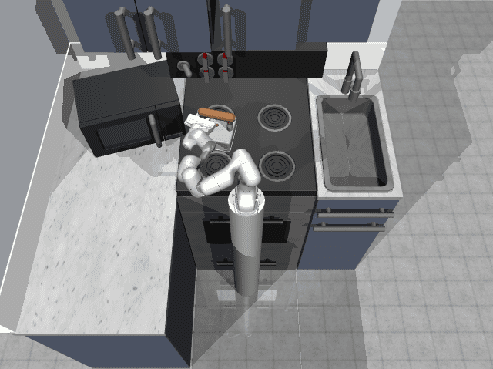}
    \includegraphics[width=0.19\textwidth]{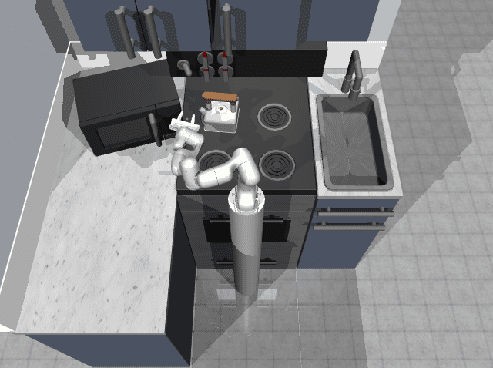}
    \includegraphics[width=0.19\textwidth]{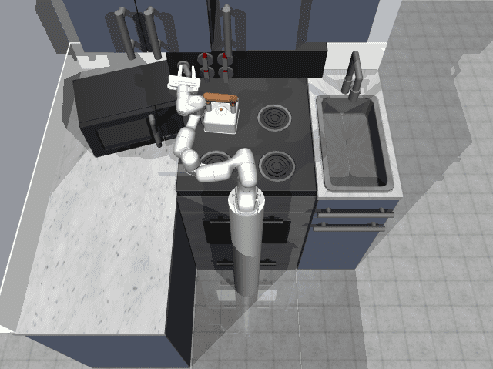}
    \includegraphics[width=0.19\textwidth]{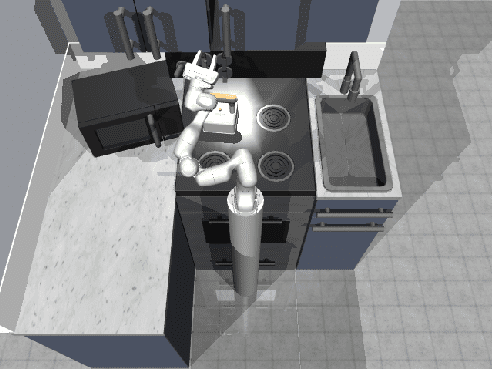}\\
    \includegraphics[width=0.19\textwidth]{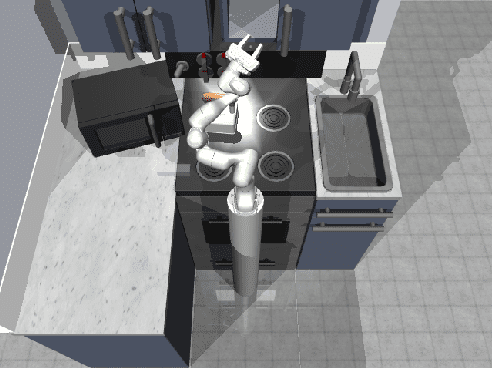}
    \includegraphics[width=0.19\textwidth]{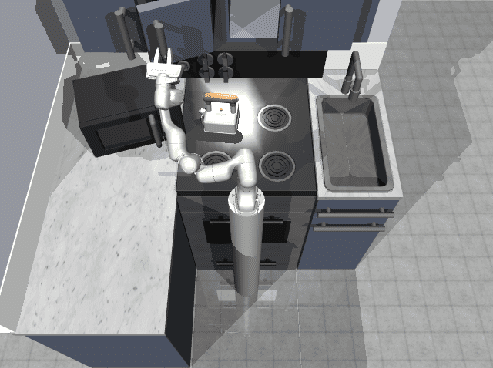}
    \includegraphics[width=0.19\textwidth]{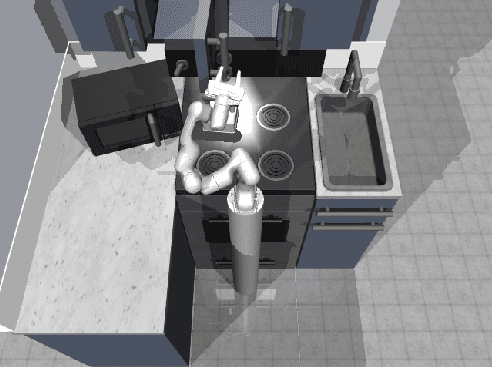}
    \includegraphics[width=0.19\textwidth]{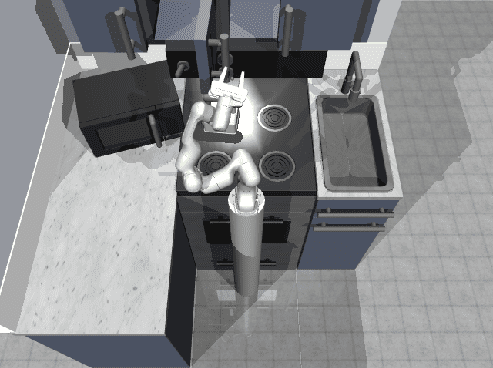}
    \includegraphics[width=0.19\textwidth]{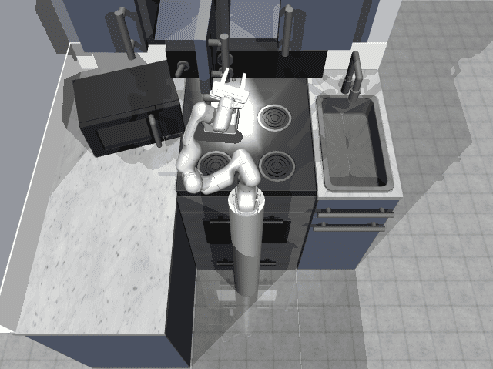}
    \caption{Visualization of successful learned behavior for moving kettle, turning top knob, sliding the slider and opening the hinge cabinet}
\end{figure}

\subsection{Failure Cases}
\begin{figure} [H]
    \centering
    \includegraphics[width=0.19\textwidth]{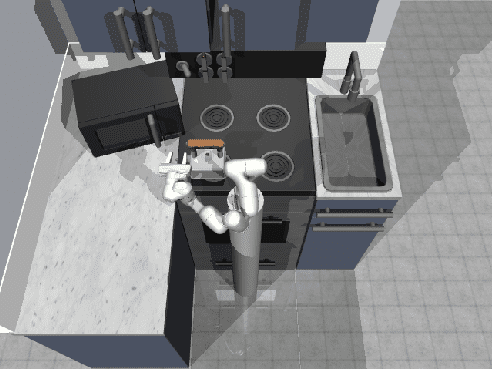}
    \includegraphics[width=0.19\textwidth]{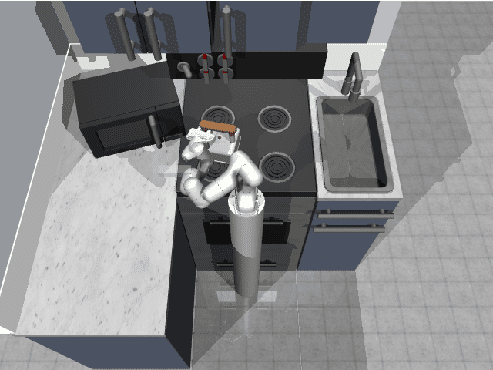}
    \includegraphics[width=0.19\textwidth]{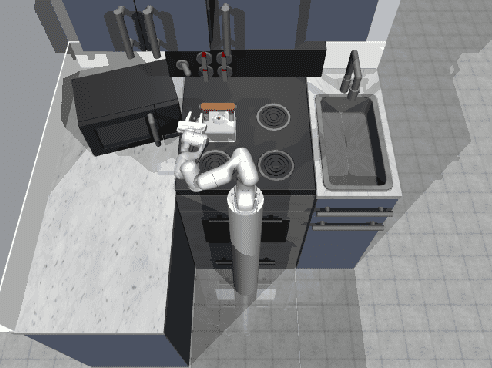}
    \includegraphics[width=0.19\textwidth]{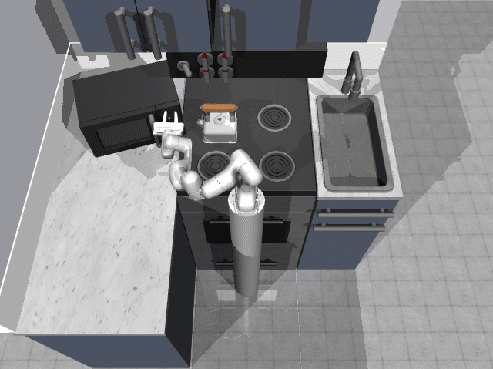}
    \includegraphics[width=0.19\textwidth]{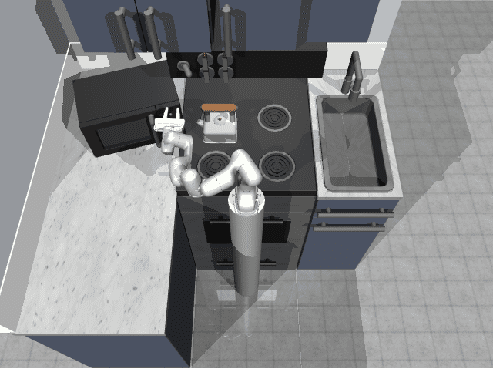}\\
    \includegraphics[width=0.19\textwidth]{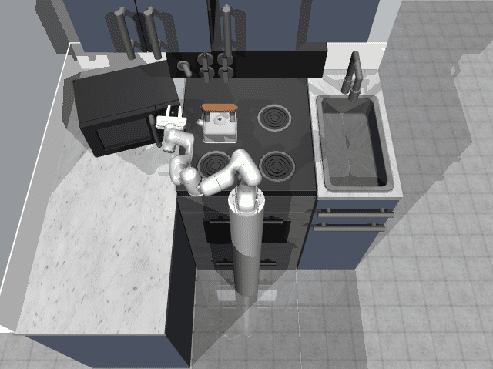}
    \includegraphics[width=0.19\textwidth]{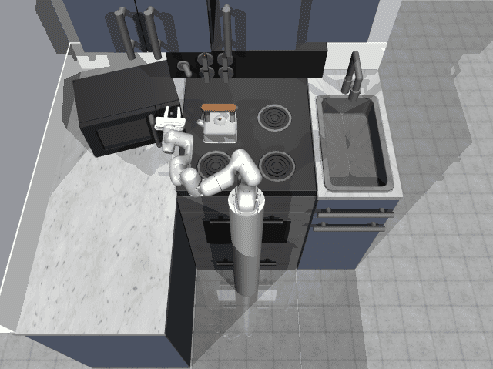}
    \includegraphics[width=0.19\textwidth]{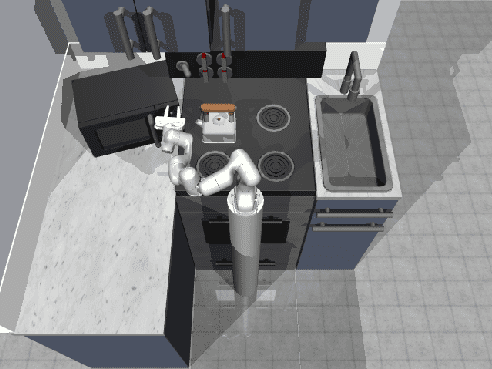}
    \includegraphics[width=0.19\textwidth]{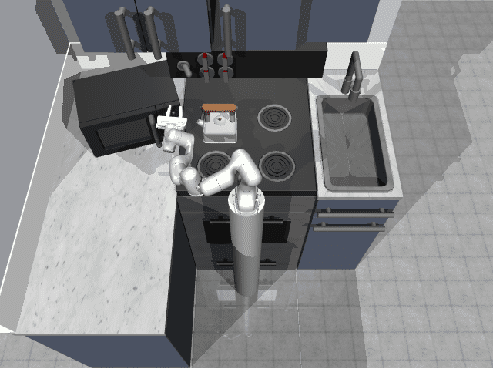}
    \includegraphics[width=0.19\textwidth]{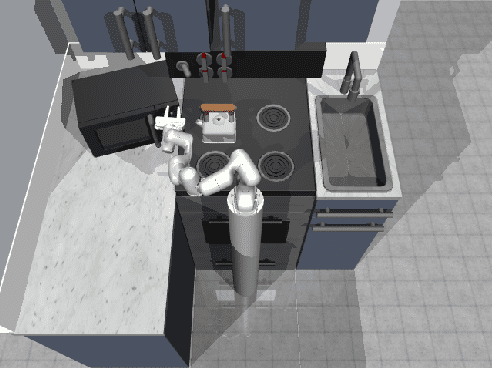}
    \caption{Visualization of failing learned behavior for moving kettle, turning the bottom knob, moving the slider and turning on the oven light}
\end{figure}

\end{document}